%% file: aaai25.tex
\documentclass[letterpaper]{article} 
\usepackage{aaai25}  
\usepackage{times}  
\usepackage{helvet}  
\usepackage{courier}  
\usepackage[hyphens]{url}  
\usepackage{graphicx} 
\usepackage{natbib}  
\usepackage{caption} 
\usepackage{algorithm}
\usepackage{algorithmic}
\usepackage{multirow}
\usepackage{color}
\usepackage{xcolor}
\usepackage{amsmath}
\usepackage{amssymb}
\usepackage{bm}
\usepackage{booktabs}
\usepackage{mathrsfs}
\usepackage{cleveref}

\usepackage{xspace}
\newcommand{\ie}{\textit{i.e.}\xspace}

\newcommand{\ourmodel}{\textit{OLiDM}}

\usepackage{newfloat}
\usepackage{listings}
\DeclareCaptionStyle{ruled}{labelfont=normalfont,labelsep=colon,strut=off} 
\floatstyle{ruled}
\newfloat{listing}{tb}{lst}{}
\floatname{listing}{Listing}
\title{\ourmodel: Object-aware LiDAR Diffusion Models for Autonomous Driving}
\author{
    Tianyi Yan\textsuperscript{\rm 1,\rm 2}\equalcontrib,
    Junbo Yin\textsuperscript{\rm 3}\equalcontrib,
    Xianpeng Lang\textsuperscript{\rm 2},
    Ruigang Yang\textsuperscript{\rm 4},
    Cheng-Zhong Xu\textsuperscript{\rm 1},
    Jianbing Shen\textsuperscript{\rm 1}\thanks{Corresponding author: \textit{Jianbing Shen}. This work was supported in part by the FDCT grants 0102/2023/RIA2 and 0154/2022/A3,
    the MYRG-CRG2022-00013-IOTSC-ICI grant, 
    and the SRG2022-00023-IOTSC grant.
    }
}
\affiliations{
    \textsuperscript{\rm 1}SKL-IOTSC, Computer and Information Science, University of Macau \quad
    \textsuperscript{\rm 2}Li Auto Inc\\
    \textsuperscript{\rm 3}CEMSE Division, King Abdullah University of Science and Technology \quad \textsuperscript{\rm 4}Shanghai Jiao Tong University\\
    \{tianyi.yan123, yinjunbocn\}@gmail.com \quad jianbingshen@um.edu.mo
}

\usepackage{bibentry}

\begin{document}

\maketitle
\begin{figure*}[t]
  \centering
  \includegraphics[width=0.88\linewidth]{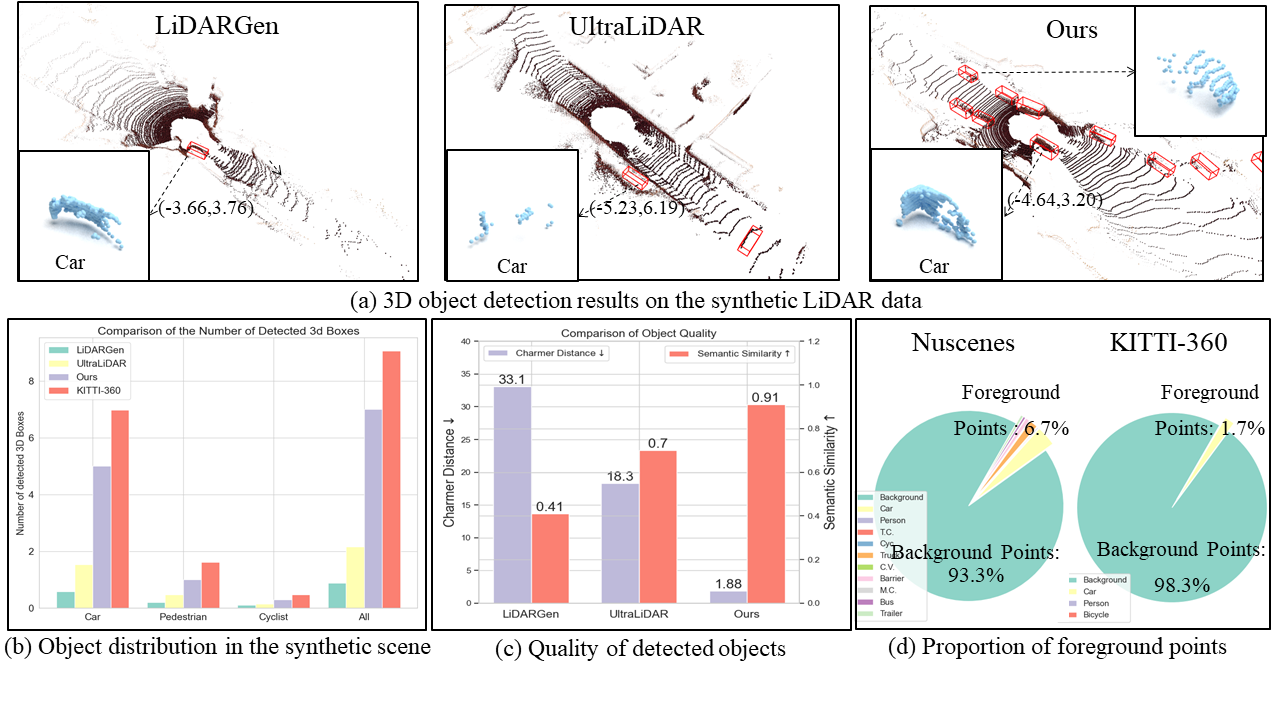}
        \vspace{-8mm}
  \caption{
  To assess the quality of foreground objects, we utilize an off-the-shelf 3D detector (\textit{i.e.}, SECOND~\cite{yan2018second} trained on KITTI~\cite{geiger2013kitti}) to identify 3D objects in LiDAR data generated by various methods, including LiDARGen, UltraLiDAR and our \ourmodel.
  \textbf{(a)} Visualization of some detected 3D objects in generated LiDAR scenes, where \ourmodel~enables the creation of LiDAR data with high-fidelity foreground objects.
 \textbf{(b-c)} We count the number and evaluate the quality of the detected objects, revealing our generated LiDAR data presents a similar distribution compared to the real data in KITTI-360~\cite{liao2022kitti360}. In contrast, LiDARGen and UltraLiDAR produce significantly fewer foreground objects with lower quality.
\textbf{(d)} Foreground object points represent a minimal fraction of the total scene points, highlighting the challenges of generating high-quality foreground objects. Please zoom in for detailed visualization.
}
  \label{Fig:motivation}
\vspace{-5mm}
\end{figure*}

\begin{abstract}
To enhance autonomous driving safety in complex scenarios, various methods have been proposed to simulate LiDAR point cloud data. Nevertheless, these methods often face challenges in producing high-quality, diverse, and controllable foreground objects. To address the needs of object-aware tasks in 3D perception, we introduce \ourmodel, a novel framework capable of generating high-fidelity LiDAR data at both the object and the scene levels.
\ourmodel~consists of two pivotal components: the Object-Scene Progressive Generation (OPG) module and the Object Semantic Alignment (OSA) module.
OPG adapts to user-specific prompts to generate desired foreground objects, which are subsequently employed as conditions in scene generation, ensuring controllable outputs at both the object and scene levels. This also facilitates the association of user-defined object-level annotations with the generated LiDAR scenes. Moreover, OSA aims to rectify the misalignment between foreground objects and background scenes, enhancing the overall quality of the generated objects.
The broad effectiveness of \ourmodel~is demonstrated across various LiDAR generation tasks, as well as in 3D perception tasks. Specifically, on the KITTI-360 dataset, \ourmodel~surpasses prior state-of-the-art methods such as UltraLiDAR by 17.5 in FPD.
%
Additionally, in sparse-to-dense LiDAR completion, \ourmodel~achieves a significant improvement over LiDARGen, with a 57.47\% increase in semantic IoU. Moreover, \ourmodel~enhances the performance of mainstream 3D detectors by 2.4\% in mAP and 1.9\% in NDS, underscoring its potential in advancing object-aware 3D tasks. Code is available at: \url{https://yanty123.github.io/OLiDM/}.
\end{abstract}

%

\section{Introduction}
\label{sec:intro}
LiDAR sensors are crucial for autonomous driving, as they can provide precise 3D information of the surroundings across various challenging conditions~\cite{li2024div2x,li2023lwsis,yan2018second,cheng2023language,yin2024fusion}. 
However, acquiring and annotating high-quality LiDAR data is costly and labor-intensive due to the sparse and noisy nature of LiDAR point cloud~\cite{meng2020anno2,wu2020deepanno1}. This highlights the necessity for advanced generation algorithms to produce diverse, controllable, and scalable LiDAR point cloud data.

Recent breakthroughs in 3D generative models \cite{lidargen,nakashima2023r2dm} have enabled promising applications such as LiDAR data generation, which are critical for self-driving vehicles. 
To generate novel LiDAR data from scratch, LiDARGen~\cite{lidargen} and R2DM~\cite{nakashima2023r2dm} transform 3D point clouds into structured 2D range images and leverage existing 2D diffusion models~\cite{ho2020denoising}. Later, UltraLiDAR~\cite{UltraLiDAR} utilizes voxelized LiDAR point clouds in conjunction with VQ-VAE~\cite{yu2021vector} to facilitate the densification of sparse LiDAR data.
Despite their effectiveness in generating scene-level LiDAR data, we have observed significant discrepancies between these synthetic data and real-world data when applied to the 3D detection task. Firstly, as shown in \cref{Fig:motivation} (a) and (c), the quality of the generated objects is relatively low. In \cref{Fig:motivation} (b), significantly fewer foreground objects are detected in these synthetic data, highlighting the distribution discrepancy with real-world data. 
In practical applications, the focus is typically on foreground objects~\cite{lang2019pointpillars}, yet existing models often generate LiDAR data without distinguishing between foreground and background, treating the scene as a whole. Additionally, foreground points are generally more sparse as shown in \cref{Fig:motivation} (d)). The challenge of developing a generative model capable of producing LiDAR data with high-quality 3D foreground objects still remains unresolved.

To address the above problems, we propose \ourmodel, an \textbf{O}bject-aware \textbf{Li}DAR \textbf{D}iffusion \textbf{M}odel that aims to produce controllable and realistic LiDAR point cloud data \textit{at both object and scene levels}.
\ourmodel~comprises two key modules: the Object-Scene Progressive Generation (OPG) module and the Object Semantic Alignment (OSA) module. 
In particular, OPG advances the generation process by progressively modeling the foreground objects and background scenes. 
To effectively model the objects, OPG incorporates rich conditions, such as detailed textual descriptions and precise geometric specifications. These conditions provide both semantic and geometric context, explicitly enhancing the diversity of objects. Then, an object denoiser receives these conditions and generates high-quality 3D objects. Subsequently, these objects serve as a prior condition to assist the scene generation process. Here, a scene controller is introduced to embed the object condition into features, enforcing the scene denoiser to produce more meaningful LiDAR scenes with controllable 3D objects. Moreover, unlike previous approaches that generate simulated LiDAR data without specific 3D object information, OPG provides initial 3D annotations for the foreground objects, which can potentially benefit the downstream 3D perception tasks.

For generating scene-level LiDAR data, previous methods \cite{nakashima2023r2dm,lidargen} typically operate on spatially disordered range images, where range values of foreground and background can be mixed in a local window. In contrast, our OSA module tactfully partitions regions into various semantic subspaces. These semantic subspaces, defined by object category, serve as semantic masks to balance the diffusion loss, thereby facilitating model learning and refining object details within the scene.
Through a meticulously designed learning process, \ourmodel~emerges as a highly practical and scalable LiDAR generator at both object and scene levels. 

Our main contributions are summarized as follows:
\begin{itemize}
    \item We introduce a new diffusion framework, \ourmodel, to ensure the generation of high-quality LiDAR objects. To the best of our knowledge, \ourmodel~is the first effort that handles controllable and realistic LiDAR data {at both object and scene levels}.
    \item We propose an Object-Scene Progressive Generation (OPG) process. OPG integrates semantic and geometric conditions to achieve precise 3D object modeling. A new scene controller is also introduced to smoothly incorporate these 3D objects into the overall LiDAR scene.
    \item An Object Semantic Alignment (OSA) module is proposed to enforce the consistency optimization of foreground objects over their respective semantic subspaces during the diffusion process, which improves the overall quality of the generated LiDAR data.
    
\end{itemize}
Extensive experiments demonstrate that \ourmodel~generates high-quality LiDAR point clouds, achieving the best FPD and JSD on KITTI-360. Moreover, we are the first to evaluate the quality of foreground LiDAR objects generated by various methods, where \ourmodel~showing superior performance across all metrics. Additionally, \ourmodel~excels in conditional generation tasks such as sparse-to-dense LiDAR completion. Finally, our validation on the nuScenes dataset confirms that \ourmodel~effectively enhances the performance of downstream 3D perception tasks, \textit{e.g.}, improving mAP by 2.4\% of mainstream 3D detectors.

\section{Related Work}

\subsection{Generative Models}
Building upon DDPM's foundation~\cite{ho2020denoising}, subsequent research \cite{li2023gligencondition, lee2021priorgradcondition, lugmayr2022repaintinpaint, saharia2022photorealistic} has explored integrating various forms of control into these models, demonstrating remarkable versatility in applications such as text-to-image synthesis and inpainting.
Shifting the focus to the 3D domain, initial research in point cloud generation focuses on Generative Adversarial Networks (GANs), with r-GAN~\cite{achlioptas2018learninggan} setting the groundwork. Subsequent enhancements \cite{zhang2022multigan,shu20193dgan} refined these models,
while recent developments \cite{cheng2023sdfusiondiffusion,zhou20213ddiffu} incorporate diffusion processes into point cloud generation. 
DiT-3D~\cite{mo2024dit3d} utilizes a diffusion transformer architecture for denoising voxelized point clouds. These advancements were primarily trained on comprehensive datasets with specific categories~\cite{chang2015shapenet,sunmodelnet40}.
Point-E~\cite{nichol2022pointe} represents a significant advancement in creating diverse 3D point clouds, leveraging large-scale text and 3D model pairings for training.
Unlike general 3D object generation, this paper tackles the more challenging scenarios in autonomous driving, addressing issues like sparsity, occlusion, and uneven point distribution. Through tactful model design, \ourmodel~effectively overcomes these obstacles, ensuring the generation of realistic and controllable LiDAR objects and scenes.
\vspace{-0.5em}
\subsection{LiDAR Data Generation}
For the task of LiDAR data generation in autonomous driving, deep generative models such as GANs~\cite{achlioptas2018learninggan,caccia2019lidargan,sauer2021projectedgan}, VAEs~\cite{caccia2019lidarvae} and their innovative hybrids have demonstrated significant advancements.
Most recently, LiDARGen~\cite{lidargen} proposes a new diffusion-based model for LiDAR data generation by transforming the input as the range image. R2DM~\cite{nakashima2023r2dm} further applies the DDPM to enhance the quality of generation. UltraLiDAR~\cite{UltraLiDAR} leverages voxelized LiDAR point clouds in conjunction with VQ-VAE~\cite{yu2021vector}, promoting efficient LiDAR data generation and completion. 
LiDM~\cite{ran2024lidm} leverages multi-modal conditions as input for LiDAR generation.
Notably, these methods mainly focus on mimicking the data distribution of entire LiDAR scenes, where background areas inevitably constitute most of the points. As a result, they overlook the inherent differences between foreground objects and background areas, compromising the overall quality. Our work shifts more attention to foreground objects, which are practical and crucial in autonomous driving. By integrating detailed conditions into a progressive generation framework from object to scene levels, \ourmodel~provides a more effective solution for LiDAR data generation.

\section{The Proposed \ourmodel~Framework}
\begin{figure*}[t]
  \centering
  \includegraphics[width=0.88\linewidth]{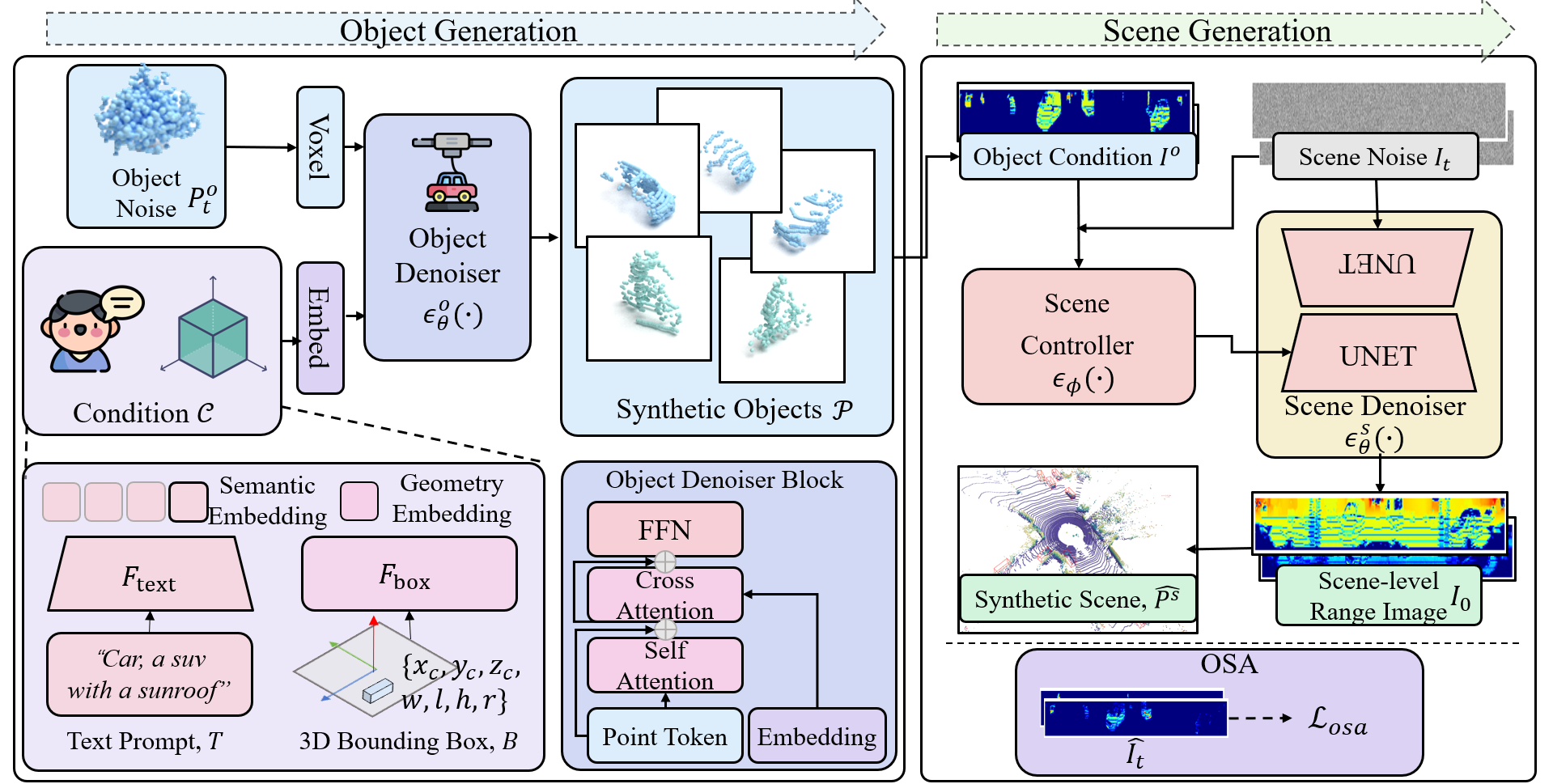}
        \vspace{-2mm}
  \caption{\textbf{Pipeline of \ourmodel.} \ourmodel~is designed to generate diverse, controllable, and realistic LiDAR point clouds at both object and scene levels through the Object-Scene Progressive Generation (OPG) process. \textbf{Object Generation:} OPG carefully combines conditions such as text descriptions and 3D geometric context to accurately model LiDAR objects. \textbf{Scene Generation:} OPG then incorporates these generated objects as specific conditions during scene-level generation, supported by a scene controller and an object semantic alignment module. 
  }
       \vspace{-1.7em}
  \label{fig:overview}
\end{figure*}

\subsection{Overall Architecture}
Formally, \ourmodel~(\textbf{O}bject-aware \textbf{Li}DAR \textbf{D}iffusion \textbf{M}odels) aims at generating both object-level point cloud $\hat{{P^o}} \in \mathbb{R}^{N^0 \times 4}$ and scene-level one $\hat{P^s} \in \mathbb{R}^{N^s \times 4}$, with input conditions $\mathcal{C}=\left\{T, B \right\}$, where $N^o$ and $N^s$ represent the maximum numbers of points, and $\hat{{P^o}}$ and $\hat{P^s}$ are characterized by 4-d attributes (\ie, the 3D coordinates $(x, y, z)$ and the LiDAR intensity $i$). $T$ denotes textual descriptions of objects and $B$ outlines their 3D bounding boxes to specify positions and orientations within the scene.
This involves an progressive object-to-scene generation process, as shown in \cref{fig:overview}: 

\textbf{\textit{Object-level}}: Given the condition
$\mathcal{C}$ and the noised point cloud $P^o_t$, an object denoiser $\epsilon_{\theta}^{o}(\cdot)$ is designed to iteratively estimate and remove the object noise $\epsilon^o$, where $\epsilon^o \sim \mathcal{N}(0,1)$ is applied on the real object-level point cloud $P^o_0 \in \mathbb{R}^{N\times4}$ to obtain $P^o_t \in \mathbb{R}^{N\times4}$ at timestep $t$. 
The intuition behind $\mathcal{C}$ is that the semantic conditions $T$ can potentially handle long-tail classes at a semantic level, while geometric conditions $B$ enhance the simulation of distant cars and rare poses at the instance level.
Accordingly, the object denoiser $\epsilon_{\theta}^{o}(\cdot)$ gradually refines $P^o_t$ towards $P^o_0$, yielding a series of synthetic LiDAR objects $\mathcal{P}=\left \{\hat{{P^o}} \right\}$ given various $\mathcal{C}$.

\textbf{\textit{Scene-level}}: 
%
The above synthetic point clouds $\mathcal{P}$ captures the details of the foreground objects, which naturally serves as a foundation for inpainting a complete LiDAR scene $P^s \in \mathbb{R}^{N^s \times 4}$. Here, we formulate $P^s$ as a range image $I \in [0,1]^{H \times W \times 2}$ to ensure computation efficiency, where $H$ and $W$ are the sensor's resolution in azimuth and elevation, respectively. 2-d channels represent distance and intensity.
To achieve high-fidelity scene-level generation, a new scene controller $\epsilon_{\phi}(\cdot)$ is introduced to assist the scene denoiser $\epsilon_{\theta}^{s}(\cdot)$ in refining the noisy scene input $I_t$ at timestamp $t$. This is accomplished by carefully encoding the object-level information $I^o$, which is extracted from $\mathcal{P}$. The scene-level and object-level latent features effectively interact to produce a more accurate and coherent representation of the scene $I_0$, \ie, enriched by high-quality 3D objects.

\subsection{Object-Scene Progressive Generation Process}
\label{sec:cpg}
Previous LiDAR generation approaches~\cite{nakashima2023r2dm,UltraLiDAR} primarily generate point cloud at the \textit{entire scene} level, overlooking the differences between foreground and background and thus inevitably degrading the quality of foreground objects. The rationale for handling the foreground points includes:
(1)~\textit{Object-Scene Point Imbalance.} The foreground typically constitutes less than 10\% of the points in an entire scene (see \cref{Fig:motivation} (d)), yet existing methods often fail to adequately address these sparse foreground points. As a result, the model may become biased toward the data distribution of background points.
(2)~\textit{Object-Scene Semantic Discrepancy.} Foreground objects often include pedestrians and vehicles, which are more critical in autonomous driving, while background areas typically consist of roads, buildings, and vegetation. Consequently, points from the foreground and background exhibit distinct geometries that need to be addressed differently.
To this end, we introduce the OPG, which employs a progressive framework—object-first, scene-later—to effectively address the inherent discrepancies between the two.
\vspace{-1mm}
\subsubsection{Object-level LiDAR Generation}
\label{sec:obj}
Unlike the point cloud objects in 3D Shapenet~\cite{chang2015shapenet} that present relatively complete and smooth patterns, 3D LiDAR objects exhibit significantly different characteristics depending on their position relative to the ego-vehicle, with distant objects often appearing sparser. As such, it is necessary to redesign current diffusion models for object-level LiDAR generation.

Given a real object point cloud \( P_{0}^{o} \in \mathbb{R}^{N \times 3} \) (omitting intensity here for simplicity) from the database, where \( N \) is the number of points with \( x, y, z \) coordinates, we first add noise $\epsilon^o$ on $P^o_0$ to obtain the noisy version $P^o_t\in \mathbb{R}^{N \times 3}$, which mainly follows DDPM~\cite{ho2020denoising}. $P^o_t$ is then voxelized and embedded as $\bm{V}_t$ by leveraging 3D convolution operations~\cite{mo2024dit3d}. 
Then, we define the conditions $\mathcal{C}=\left\{T,B\right\}$. 
Specifically, $T$ is formulated as ``\textit{An object from class} {$\langle$category$\rangle$}, {$\langle$description$\rangle$}." 
Here, $\langle$category$\rangle$ refers to the object's category, as annotated in the datasets, while $\langle$description$\rangle$ offers detailed instance-level descriptions, such as ``\textit{a sports car that is on the street, side view}". During training, we employ an advanced caption model~\cite{li2022blip} to provide rich descriptions for each 3D object. Then, we get the embeddings of $T$ by:
\begin{equation}
    [\bm{f}^T_\text{CLS}, \bm{f}_{0}^T, \cdots, \bm{f}_{L}^T]=F_\text{text}({T}),
\end{equation}
where $L$ denotes the length of the sentence, $F_\text{text}(\cdot)$ is the pre-trained CLIP~\cite{chen2020simpleCLIP}.
After that, we let $\bm{f}^{T}=\bm{f}^T_\text{CLS}$ to summarize $T$ since $\bm{f}^T_\text{CLS}$ has effectively integrated the context by attending to all positions in $[\bm{f}^T_\text{CLS}, \bm{f}_{0}^T, \cdots, \bm{f}_{L}^T]$.
Meanwhile, 
we propose to utilize the 3D bounding box, $B=[x_c,y_c,z_c,w,l,h,r]$, to provide crucial object geometric information that indicates the distribution of points. To be specific, the 3D center and rotation reveal the distance and orientation relative to the ego-vehicle, which in turn determines the visibility of the object's points. Additionally, the dimensions of the box potentially offer a unique identifier for each instance within the same category.
Here, we extract geometry-aware embeddings $\bm{f}^{B}$ by:
\begin{equation}
    \bm{f}^{B}=F_\text{box}(B)=F_\text{fe}([x_c,y_c,z_c,w,l,h,r]),
    \label{eq:box}
\end{equation}
where $F_\text{fe}(\cdot)$ denotes Fourier Embedder \cite{mildenhall2021nerf}.
Furthermore, $\bm{f^T}$ and $\bm{f^B}$ are combined to obtain a comprehensive condition embedding $\bm{c}$:
\begin{equation}
    \bm{c}=F_\text{com}([\bm{f^B},\bm{f^T}]),
\end{equation}
where $[\cdot, \cdot]$ is the concatenation operation and $F_\text{com}(\cdot)$ can be realized by linear layers.

Since we have the noised input embedding $\bm{V}_t$ and the condition embedding $\bm{c}$, an Object Denoiser (See \cref{fig:overview}) is proposed to estimate the noise at timestep $t$. Inspired by existing diffusion models~\cite{nichol2022pointe,luo2021diffusionpoint}, we apply the cross-attention mechanism to enhance the interaction between the inputs. The query, key and value in the cross-attention layer are denoted as:
\begin{equation}
    \textbf{q}=\text{linear}(\bm{V}_t) ,
    \textbf{k}=\text{linear}(\bm{c}+\bm{f^t}) ,
    \textbf{v}=\text{linear}(\bm{c}+\bm{f^t}),
\end{equation}
where $\bm{f^t}$ is the embedded timestep $t$. In this way, the noised voxel embedding $\bm{V}_t$ can integrate information from $\bm{f^t}$ and $\bm{c}$ to refine its 3D representation:
\begin{equation}
    \bm{\hat{V}_t}=\bm{V_{t}}+\textbf{v}^{\mathsf{T}} \cdot \text{softmax}(\textbf{k} \cdot \textbf{q}^{\mathsf{T}})
\end{equation}
Finally, a feed-forward network estimates an 
$N \times 3$ tensor based on the attention output $\bm{\hat{V}_t}$, representing the object noise towards the ground truth $\epsilon^o$. The Object Denoiser $\epsilon_{\theta}^{o}(\cdot)$ can be optimized as:
\begin{equation}
        \mathcal{L}_\text{object}=\mathbb{E}_{{P}^{o}_{0},\bm{c},\epsilon^o,t}[\left \| \epsilon^o - \epsilon_{\theta}^{o} (P^o_t, t, \bm{c}) \right \|^2].
    \label{eq:dif}
\end{equation}

During inference, the Object Denoiser iteratively processes the noised input object points with condition to generate a synthetic 3D object. This can be repeated to produce an unlimited number of point cloud objects $\mathcal{P}$ based on different $\mathcal{C}$. Notably, users have the flexibility to define semantic and geometric conditions to account for different object distribution settings. The process can be simplified by applying uniform or random sampling for $B$ on the road and specifying only the category description for each object as $T$. 

\subsubsection{Scene-level LiDAR Generation}
Building upon the object point clouds $\mathcal{P}$, our goal is to generate complete LiDAR scenes $P^s$ that are compatible with these high-quality objects. The main challenge in scene-level point cloud generation lies in handling hundreds of thousands of points simultaneously. Following the methodologies in \cite{lidargen}, we choose to represent the entire LiDAR scene as a range image to save computation. 

Specifically, object points $\mathcal{P}$ are first projected into an object-level range image \( I^o \). Then, a scene controller $\epsilon_{\phi}(\cdot)$ is introduced to maintain the contents and positions of the objects and guide the generation process. This is achieved by initially extracting a conditioning embedding and combining it with the noised scene input $I_t$ to produce the intermediate latent feature $\bm{h_c^t}$:
\begin{equation}
\bm{h_c^t} = \text{zero}(\text{conv}(I^o)) + \text{zero}(\text{conv}(I_t))
\end{equation}
Here, $\text{zero}$ denotes the zero convolution layer \cite{zhang2023controllnet}. Following several layers of stacked U-Net blocks \cite{ronneberger2015unet} and zero convolutions, we refine $\bm{h_c^t}$ and effectively control the detailed rendering of foreground objects.

The scene denoiser $\epsilon_{\theta}^{s}(\cdot)$ then processes the scene noise $I_t$ along with the control embedding to extract the final latent feature. The optimization process of the above process can be denoted as: 
\begin{equation}
        \mathcal{L}_\text{object-scene}=\mathbb{E}_{I^o,{I}_{0},\epsilon^s,t}[\left \| \epsilon^s - \epsilon_{\theta}^{s} (I_t, t,\epsilon_{\phi}(I^o)) \right \|^2].
    \label{eq:difscene}
\end{equation}
By leveraging this progressive generation framework, \ourmodel~not only ensures the realism and fidelity of individual object details but also preserves the contextual integrity and coherence of the entire scene, thereby resulting in highly realistic and contextually accurate LiDAR scenes. 

\begin{figure}[t]
  \centering
    \includegraphics[width=0.95\linewidth]{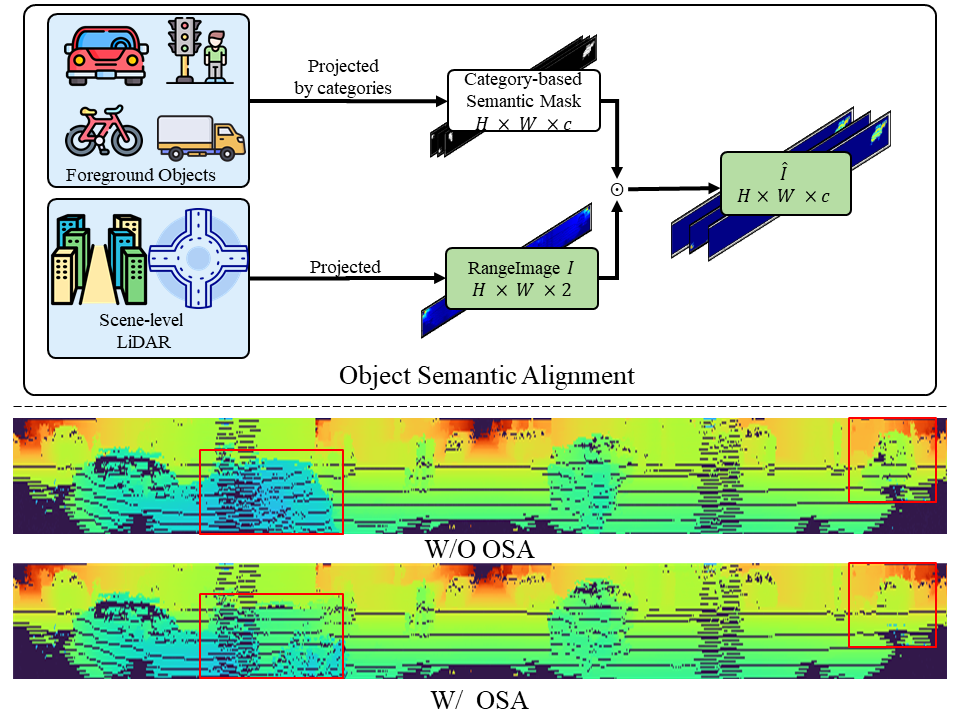}
      \vspace{-1em}
  \caption{\textbf{The Object Semantic Alignment (OSA) module} aligns object features based on their semantic space, enhancing the foreground object generation and contributing to the overall quality of the generated LiDAR scenes. 
  }
  \label{fig:condition}
    \vspace{-7mm}
\end{figure}

\subsection{Object Semantic Alignment}
\label{sec:osa}
Here, we introduce the proposed OSA module. While range images contain rich depth information, they suffer from \textit{spatial misalignment} issues \cite{bai2024rangeperception}, where adjacent pixels in the range image can represent real-world distances greater than 30 meters. Previous approaches \cite{UltraLiDAR,nakashima2023r2dm} directly feed entire range images into a 2D backbone, resulting in mixed features for near and distant objects. This hinders the extraction of accurate geometric information of the foreground objects and corrupts the generation quality.

To rectify the object-specific representations from the mixed features, we design the OSA module, which aligns features within the semantic subspace for each 3D object. By splitting the features into distinct spaces based on categories, we minimize interference between regions. As illustrated in \cref{fig:condition}, this method sharpens edge information for foreground objects, resulting in the higher-quality generation.

Specifically,
given a frame of range image $I\in \mathbb{R}^{H \times W \times 2}$, OSA first calculates binary mask $M \in \mathbb{R}^{H \times W \times c}$ according to foreground objects $\mathcal{P}$, where $c$ is the number of categories readily defined in the object generation process, and each $M_i \in {H \times W}$ is a pixel-wise mask for range image $I$, indicating whether each point represents the object from $c_i$ category.
Next, OSA defines a tensor $\hat{I} \in \mathbb{R}^{H \times W \times c}$, which is a $c$-channel version of $I$, with the $i$-th channel defined by $\hat{I}_i=I \odot M_i$. Here, $\hat{I}_t$, the noisy version of $\hat{I}$ serves as input of the scene denoiser $\epsilon^s(\cdot)$ to calculate OSA loss:
\begin{equation}
        \mathcal{L}_{\text{osa}}=\mathbb{E}_{\hat{{I}_{0}},\hat{\epsilon^s},t}[\left \| \hat{\epsilon^s} - \epsilon^s_{\theta} (\hat{I_t}, t) \right \|^2],
    \label{eq:ssa}
\end{equation}
where $\hat{\epsilon^s} \in \mathbb{R}^{H \times W \times c}$ is $t$ steps noise we added on $\hat{I}$. 
We adopt \cref{eq:ssa} as an additional loss with \cref{eq:difscene} to achieve alignment on the semantic feature subspace at the category level, thereby enhancing the quality of foreground object generation and ensuring a more consistent integration between foreground and background.

\begin{figure*}[t]
  \centering
  \includegraphics[width=0.80\linewidth]{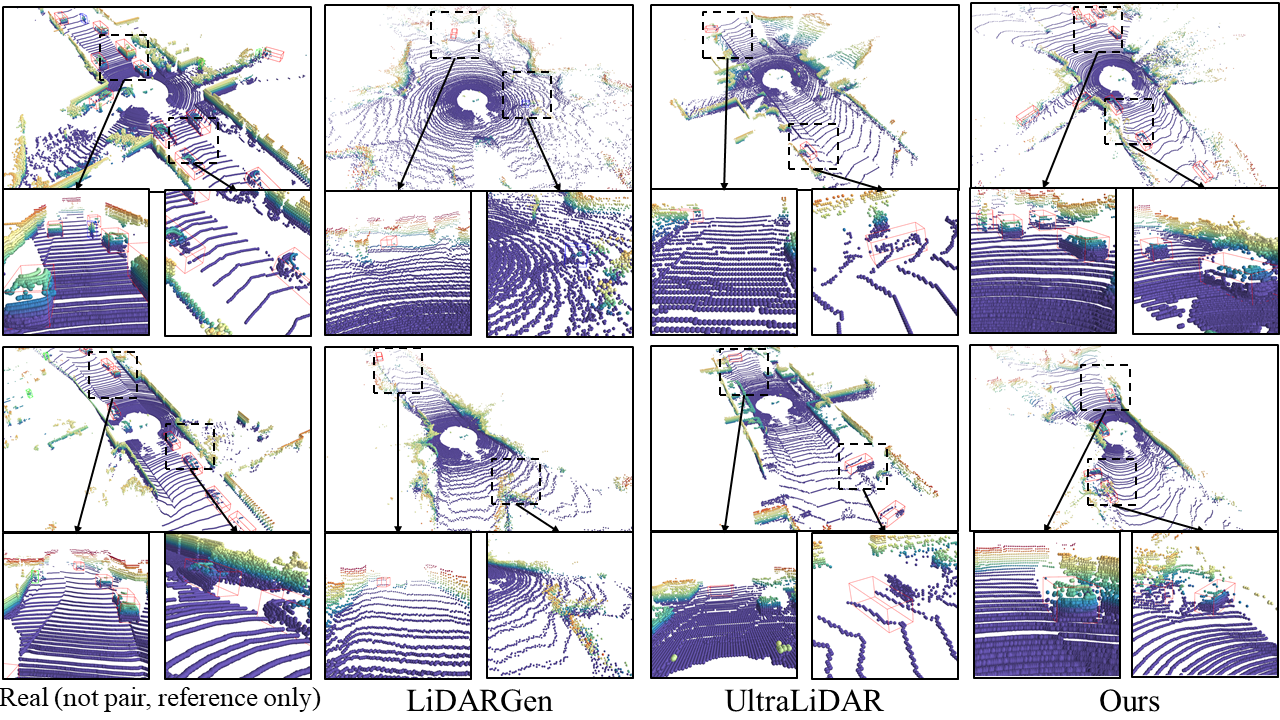}
  \vspace{-2mm}
  \caption{\textbf{Qualitative comparison against baselines on LiDAR generation.} We compare with LiDARGen, UltraLiDAR and include real LiDAR for reference. \ourmodel~generates LiDAR data with more realistic sparsity and beam patterns. \textcolor{red}{Red}, \textcolor{blue}{Blue} and \textcolor{green}{Green} boxes are the detected objects (car, cyclist and pedestrian) from a 3D detector trained on KITTI.}
  \vspace{-3mm}
  \label{fig:vis}
\end{figure*}

\section{Experiments}
\subsection{Experimental Setups}
\textbf{Dataset and Setting.}
nuScenes~\cite{caesar2020nuscenes} and KITTI-360~\cite{liao2022kitti360} are popular datasets widely used in autonomous driving research, featuring detailed annotations for evaluating different tasks.
For the generation task, we employ the KITTI-360 dataset to demonstrate the effectiveness, while for 3D object detection and other tasks~\cite{han2024weakly,tao2023weakly}, we use the nuScenes \cite{caesar2020nuscenes} as the benchmark.  

\noindent\textbf{Evaluation Metrics.}
For the scene-level LiDAR generation, we use MMD, JSD on the BEV plane and FPD as the metrics, and we generate 10k samples using 0000 and 0002 seq as the validation split following \cite{lidargen,UltraLiDAR}.
For the quality of foreground objects, we use Chamfer Distance (CD), and Jensen-Shannon Divergence (JSD) to judge. We also define Semantic Similarity (SS) to evaluate semantic closeness using cosine similarity of embeddings from the pre-trained Openshape~\cite{liu2023openshape}. 
Details are included in Section 1.2 of supplementary materials. R2DM is our baseline, with range image inputs (Sec.3.1 $w=1024$).

\begin{table}[t]
    \centering
    \begin{tabular}{c|c|c|c}
    \hline
    {Method}      &{$\text{JSD}_{\small(\downarrow)}$}&$\text{MMD}_{(\downarrow)}$ & {$\text{FPD}_{(\downarrow)}$}  \\
    \hline
    LiDARVAE & 1.61 & 10.0 & -\\
    Proj.GAN  & 8.05  & 3.47 & -\\
    LiDARGen                    &  0.67  & 3.87 & 90.3\\
    UltraLiDAR                & $0.71$ & $\textbf{1.96}$ & 25.1 \\
    R2DM               & 0.49 & $4.35$ &  14.6    \\
    \hline
    \ourmodel~w/o OSA      & 0.47 & 3.89 & 9.20\\
    \ourmodel              & \textbf{0.42} & $3.45$ & \textbf{7.60}\\
    \hline
    \end{tabular}%
        \vspace{-2mm}
    \caption{\textbf{Quantitative results of the LiDAR generation task on KITTI-360.} $\text{MMD}$ and $\text{JSD}$ are multiplied with $10^{4}$ and $10$, respectively.}
    \label{tab:sota_s}
    \vspace{-6mm}
\end{table}
\begin{table}[t]
    \centering
     \begin{tabular}{c|c|c|c|c}
    \hline
   Method & $\text{CD}_{(\downarrow)}$ &$\text{SS}_{(\uparrow)}$&$\text{JSD}_{(\downarrow)}$&{\#Box}\\
    \hline
    LiDARGen  &33.1  & 0.41       &0.97&0.88\\
    UltraLiDAR          & 18.3      & 0.70&0.95 & 2.16\\
    R2DM               & 17.6   & 0.86& 0.91  & 4.66 \\
    \hline
    $\ourmodel^{-}$          & 14.3     &0.84 & 0.93 & 5.02\\
    $\ourmodel^{T}$          & 9.21     &0.86 & 0.87 & 5.92\\
        $\ourmodel^{B}$          & 4.32     &0.85 & 0.82 & 6.77\\
    \ourmodel          & \textbf{1.88}     &\textbf{0.91} & \textbf{0.74} & \textbf{7.10}\\
     \hline
    \end{tabular}%
        \vspace{-2mm}
    \caption{\textbf{Quantitative results of foreground LiDAR objects on KITTI-360.}  We sample 1000 objects for the evaluation. $\ourmodel^{-}$ indicates training without any condition while $\ourmodel^{T}$, $\ourmodel^{B}$ indicate training with text prompt $T$ and 3D box $B$, respectively. \#Box indicates the number of detected 3D boxes.}
    \label{tab:sota_o}%
        \vspace{-8mm}
\end{table}

\subsection{Object-Aware LiDAR Generation}
\label{sec:exp}
\noindent\textbf{Scene-level Generation.}
\ourmodel~is compared against several state-of-the-art LiDAR data generation methods, including LiDARGen~\cite{lidargen}, 
R2DM~\cite{nakashima2023r2dm}, and UltraLidar~\cite{UltraLiDAR}, as detailed in \cref{tab:sota_s}. 
\ourmodel~achieves superior performance in terms of JSD and FPD metrics, indicating that our \ourmodel~produces the most realistic point clouds in both BEV and point spaces.
Although UltraLiDAR shows better outcomes on the MMD metric due to its voxel-based BEV modeling, it falls short in the FPD metric, which assesses point-level accuracy. 
\ourmodel~exceeds UltraLiDAR for 17.5 at FPD. 
The enhancements stem from \ourmodel's innovative strategy, which ensure a more faithful and robust synthesis of LiDAR data.

\noindent\textbf{Object-level Generation.}
Previous LiDAR point cloud generation methods often overlook quality control at the object level, potentially limiting their applicability in autonomous driving. In response, we propose to evaluate the quality of foreground objects within the generated LiDAR data, as demonstrated by \cref{tab:sota_o}. We utilize a well-trained 3D detector \cite{yan2018second} to identify foreground objects within LiDAR scenes. 
A critical metric in our analysis, \#Box, represents the average count of detected 3D boxes per scene, with values approaching the KITTI-360 dataset's standard of 9.06 reflecting superior performance.
\ourmodel~notably aligns with the distribution of real foreground objects, significantly reducing the CD metric to 1.88 and JSD metric to 0.74, demonstrating its effectiveness in generating high-fidelity LiDAR objects.
UltraLiDAR
struggles on KITTI-360 due to its voxelization approach, which excessively sparsifies the dense point clouds of foreground objects, resulting in a mere 2.16 detected boxes. Meanwhile, our method captures 7.1 boxes, significantly closer to the real data benchmark of 9.06.
These experiments indicate the critical role of \ourmodel~n generating high-fidelity LiDAR objects.

\noindent\textbf{Qualitative Comparison.}
The qualitative comparison is presented in \cref{fig:vis}, where LiDARGen demonstrates acceptable performance in foreground patterns but struggles with scene-level fidelity. Although UltraLiDAR shows impressive scene-level patterns through its BEV representation, it sacrifices object-level quality, probably due to the BEV representation's inability to adequately capture 3D object details. In contrast, our method maintains high quality in both background structures and foreground objects, highlighting the effectiveness of the OPG and OSA designs.
\begin{table}[t]
  \centering
    \begin{tabular}{c|c|c|c}
    \hline
    \multirow{2}{*}{Method}      &\multicolumn{1}{|c|}{Depth}&\multicolumn{1}{|c|}{Reflectence} & {Semantics}  \\
    \cline{2-4}
    &MSE$\downarrow$  &MSE$\downarrow$ & IoU{\%}$\uparrow$\\
    \hline
    Nearest-neighbor  & 2.069  & 0.100  & 20.00\\
    Bilinear          & 2.193  & 0.106  & 18.19\\
    Bicubic           & 2.314  & 0.110  & 17.26\\
    \hline
    LiDARGen         & 1.551   & 0.080  & 22.46\\
    R2DM        & 0.923   & 0.050   & 34.44\\
    \ourmodel         & \textbf{0.702}  & \textbf{0.048}  & \textbf{79.93}\\
     \hline
    \end{tabular}%
          \vspace{-2mm}
      \caption{\textbf{Quantitative results of sparse-to-dense LiDAR generation on KITTI-360,} where we input 25\% beams.
    }
          \vspace{-2mm}
  \label{tab:s2d}%
\end{table}%

\begin{table}[t]
  \centering
    \begin{tabular}{c|c|c|cc}
    \hline
    {3D Detector}&{Aug. Method}&{Paradigm} &{mAP}  & {NDS}   \\
    \hline
    \multirow{3}{*}{PointPillars}  &Baseline&None& 44.8& 58.3  \\ 
    & + GT-Aug     & Real &45.1& 58.9  \\ 
    &+LiDAR-Aug & Synthetic&45.6 &58.4  \\
    &+ \ourmodel &Synthetic & \textbf{46.7}& \textbf{61.1} \\ 
    \hline
    \multirow{4}{*}{CenterPoint} &Baseline&None& 59.0& 65.8        \\ 
    &+ GT-Aug & Synthetic&59.2& 66.5  \\
    &+ LiDAR-Aug & Synthetic&60.5& 67.3 \\
    &+ \ourmodel &Synthetic & \textbf{61.9}& \textbf{68.5} \\ 
    \hline
    \end{tabular}
      \vspace{-2mm}
    \caption{\textbf{Evaluation of the effectiveness in augmenting mainstream 3D detectors} on the nuScenes. Compared with other data augmentation techniques, \ourmodel~achieves superior performance across various 3D detectors.}
  \label{tab:detection}%
  \vspace{-4mm}
\end{table}%

\subsection{Conditional LiDAR Generation}
\begin{figure}[t]
  \centering
  \includegraphics[width=\linewidth]{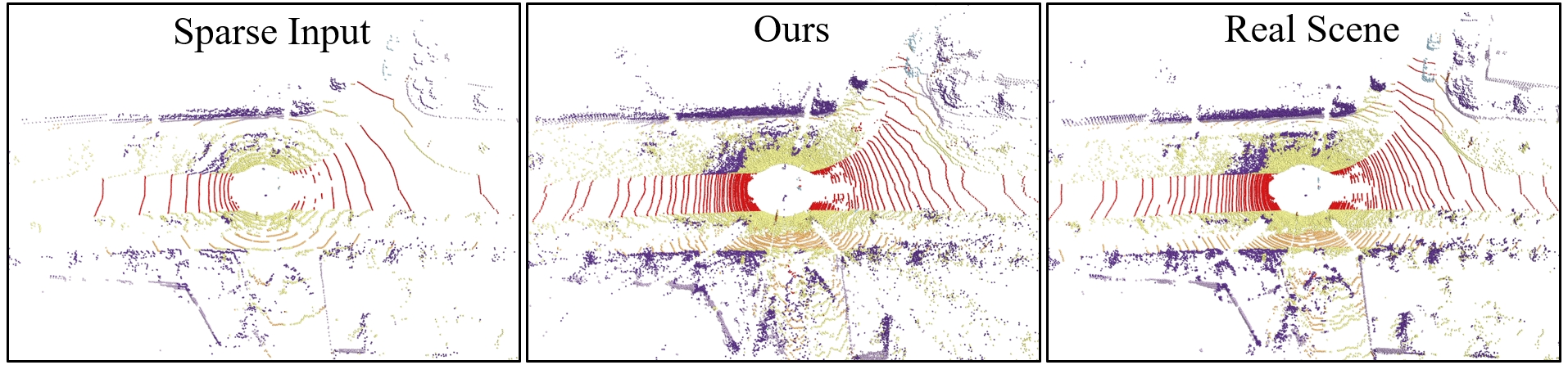}
      \vspace{-1.2em}
  \caption{\textbf{Sparse-to-dense Completion.} The semantic results are predicted by RangeNet-53 \cite{milioto2019rangenet++}.
  }
  \label{Fig:s2d}
    \vspace{-3mm}
\end{figure}

\noindent\textbf{Sparse-to-Dense LiDAR Completion.}
Following the settings of LiDARGen and R2DM, we conduct LiDAR upsampling experiments on the KITTI-360 dataset, using 16-beam LiDAR as input and producing 64-beam LiDAR as output. As shown in \cref{tab:s2d} and \cref{Fig:s2d}, \ourmodel~achieves the best results across all metrics, with a significant improvement in semantic IoU from 34\% to 79\%. The success can be attributed to the OPG's ability to stably and reliably encode conditions and facilitate feature interaction, thus controlling the overall content generation.

\begin{figure}[t]
  \centering
  \includegraphics[width=\linewidth]{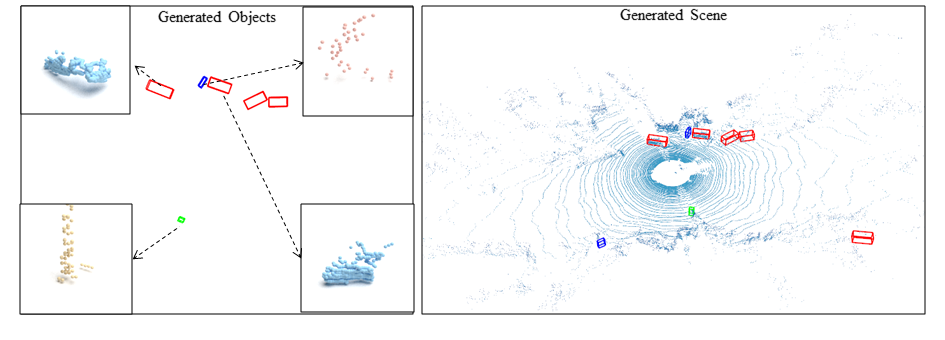}
  \vspace{-2em}
  \caption{\textbf{\ourmodel~controls contents at both object and scene levels.} 
  \ourmodel~guarantees that the desired 3D objects are accurately represented in the final LiDAR scenes.
  }
  \vspace{-4mm}
  \label{Fig:control}
\end{figure}

\noindent\textbf{Controllable LiDAR Generation.}
To verify our method's capability to control the generation of LiDAR data, we conduct validations from both object-level and scene-level control perspectives as shown in \cref{Fig:control}.
For the object level, we manipulate the textual descriptions and spatial geometric positions to control the generated object's semantic content and LiDAR scanning patterns. 
For the scene level, we leverage foreground objects as conditions for the generation of desired LiDAR scenes, including extremely crowded scenarios. This advantage demonstrates our ability to generate LiDAR data that aligns with user expectations and adheres to real-world physical rules, which is crucial for the autonomous driving industry in creating specific corner-case scenarios.

\subsection{Downstream Task}
The above experiments demonstrate that \ourmodel~can produce high-quality 3D LiDAR objects, which can be naturally leveraged to enhance downstream 3D detection task~\cite{yin2021graph}.
To further validate our approach, we employ CenterPoint~\cite{yin2021center} and Pointoillars~\cite{lang2019pointpillars} as baselines to examine improvements in detection performance. 
Specifically, we augment the detectors using the 3D objects generated by~\ourmodel, and compare with the widely used GT-Aug~\cite{yan2018second} technique (GT-Aug introduces randomly sampled ground truth from other scenes into current scene to facilitate the training). As shown in~\cref{tab:detection}, \ourmodel~improves the detector performance by 2.7\% compared to GT-Aug.
These results demonstrate that high-quality 3D objects potentially enhance the 3D perception task. It is worth noting that other current LiDAR generation methods are unable to produce controllable and high-fidelity 3D LiDAR objects, which limits their applicability in downstream perception tasks.

\section{Conclusion}
In this paper, we introduced \ourmodel, a novel framework for generating realistic and controllable LiDAR data at both object and scene levels. By leveraging the Object-Scene Progressive Generation (OPG) and Object Semantic Alignment (OSA), \ourmodel~not only captures data distribution similar to real-world LiDAR data, but also enhances the performance of downstream perception tasks. Our evaluations across diverse benchmarks highlight the model's ability to produce high-fidelity LiDAR data. We anticipate that \ourmodel~can inspire further advancements in LiDAR simulation and enhance the safety of self-driving vehicles.

\bibliography{aaai25}

\newpage
\appendix

\input{main}

\end{document}

%% file: main.tex
\section{Experimental Details}
\subsubsection{Training Dataset}
\label{sec:dataset}
Existing 3D shape datasets like ShapeNet~[\textcolor{blue}{6}] are limited in the category scale and the modality domain, posing challenges for object-level LiDAR generation. 
To enable realistic LiDAR object generation, we adopt large-scale real-world LiDAR datasets, nuScenes~[\textcolor{blue}{5}] and KITTI360~[\textcolor{blue}{19}]. 

For nuScenes dataset, the 3D LiDAR objects from diverse commonly seen categories (\ie, car, cyclist, truck, etc.) in the datasets are utilized as the learning targets. 
Based on this, we construct object-level training datasets that aligns multimodal conditions $\mathcal{C}$ with real LiDAR objects $\mathcal{P}$ to aid in training and evaluating \ourmodel. We sample 55k ground-truth LiDAR objects from nuScenes.
As for KITTI-360, we utilize a well-trained 3D detector (\ie, SECOND~[\textcolor{blue}{38}] trained on KITTI~[\textcolor{blue}{12}]) to identify foreground objects on KITTI360.
We also sample 55k real LiDAR objects from these detected objects.
\subsubsection{Evalution Metrics}
\begin{itemize}
    \item \textbf{MMD-{BEV}} and \textbf{JSD-{BEV}} with a $100 \times 100$ 2D histogram on the bird's eye view (BEV) plane are used as the metrics for the scene-level LiDAR generation. They calculate the distance between distributions of BEV occupancy grids, following~[\textcolor{blue}{47},\textcolor{blue}{28},\textcolor{blue}{37}].
    \item \textbf{Fréchet Point Cloud Distance (FPD)}~[\textcolor{blue}{47}] calculates the Fréchet distance between Gaussian distributions fitted to real and synthetic point clouds. Lower FPD values suggest higher diversity.
    \item
\textbf{Chamfer Distance (CD)}~[\textcolor{blue}{22}] measures the proximity between two point clouds by averaging the nearest neighbor distances. Lower CD indicates closer similarity. It is used for evaluating the quality of foreground objects.
\item \textbf{Semantic Similarity (SS)} is defined to evaluate semantic closeness using cosine similarity of embeddings from the pre-trained openshape model~[\textcolor{blue}{20}]. A higher SS represents a higher semantic similarity between the synthesized objects and the real objects.
\item \textbf{Mean Absolute Error (MAE)} and \textbf{Root Mean Squared Error (RMSE)} for depth and reflectance are computed for the evaluation of the sparse-to-dense LiDAR completion. We also calculate \textbf{intersection-over-union (IoU)} between the point cloud labels predicted from upsampled and real data using pre-trained RangeNet-53~[\textcolor{blue}{25}].
\end{itemize}

\subsubsection{Implement Details}
For the object-level generation, we set the input voxel size $V$ as $32$ and the number of points $N$ as 1024 following DiT-3D~[\textcolor{blue}{27}]. 
Then, we normalize both coordinates and intensity of the input to the average ranges obtained by categories on the dataset, and we also normalize the coordinates of the 3D box $B$ to $[-1,1]$ before the Fourier transforming~[\textcolor{blue}{24}].
The object denoiser is built following DiT-3D with 12 layers, and we train it on 50,000 training samples for 100 epochs using the Adam optimizer with a batch size of 128 and a learning rate of $10^{-5}$, using 4 NVIDIA A100 GPUs for roughly 12 hours. 
We crop the 2D patches corresponding the LiDAR objects and caption them with a pre-trained caption model BLIP2~[\textcolor{blue}{17}] to obtain LiDAR objects with paired textual descriptions.

For the scene-level generation, we set the height $H$ of the range image $I$ as 64 and 32 for KITTI360 and nuScenes datasets, respectively. The scene denoiser and the scene controller are optimized jointly for 300k steps with a batch size of 16 and a learning rate of $10^{-5}$. The scene denoiser is a 8 layer classical U-Net architecture~[\textcolor{blue}{28}] while the scene controller is based on the controlnet with the same architecture as the scene denoiser. 

\section{Qualitative Analysis of OSA}
\begin{figure}[t]
  \centering
  \includegraphics[width=1\linewidth]{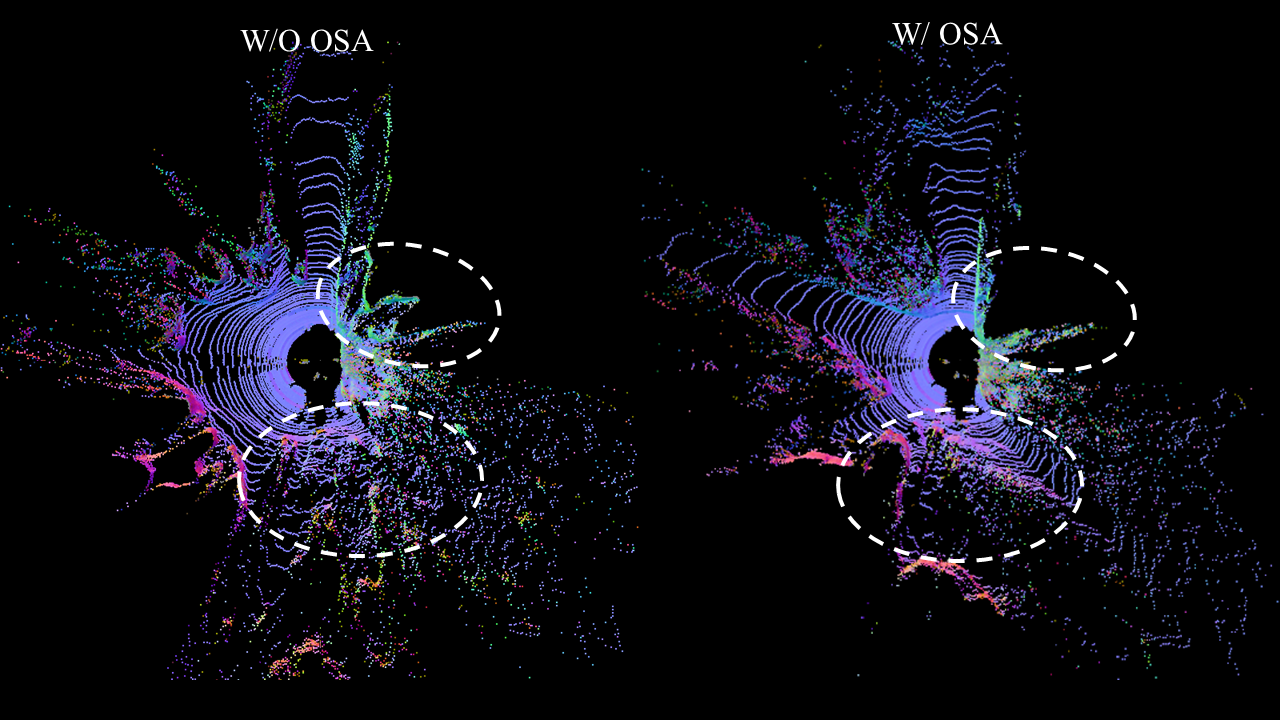}
  \caption{\textbf{Qualitative comparison between \ourmodel~w/o and w/ OSA.} OSA significantly mitigates the noise in the point cloud at the foreground-background boundary areas, making the foreground object more clearly separated from the background scene.}
  \label{fig:osa}
\end{figure}

To mitigate the noisy features resulted from spatial misalignment issues and improve the quality of LiDAR generation, we propose OSA, which aligns the features within extra channels. We illustrate qualitative results about OSA in \cref{fig:osa}. OSA significantly mitigates the noise in the point cloud at the foreground-background boundary areas, making the foreground object more clearly separated from the background scene, improving the quality of generated LiDAR data.
\begin{figure}[t]
  \centering
  \includegraphics[width=1\linewidth]{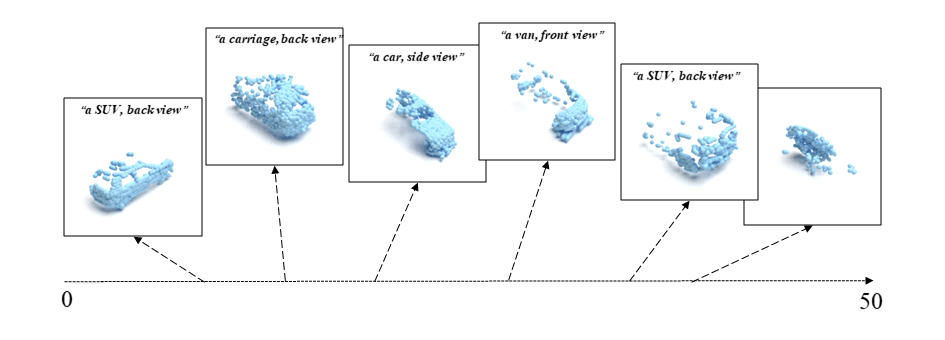}
  \caption{ \textbf{Qualitative results about object-level generation on KITTI360.} \ourmodel~controls the generation of LiDAR objects with different 3D location and text prompts. The farther away an object is, the sparser the point clouds are.}
  \label{fig:location1}
\end{figure}
\begin{figure}[t]
  \centering
  \includegraphics[width=1\linewidth]{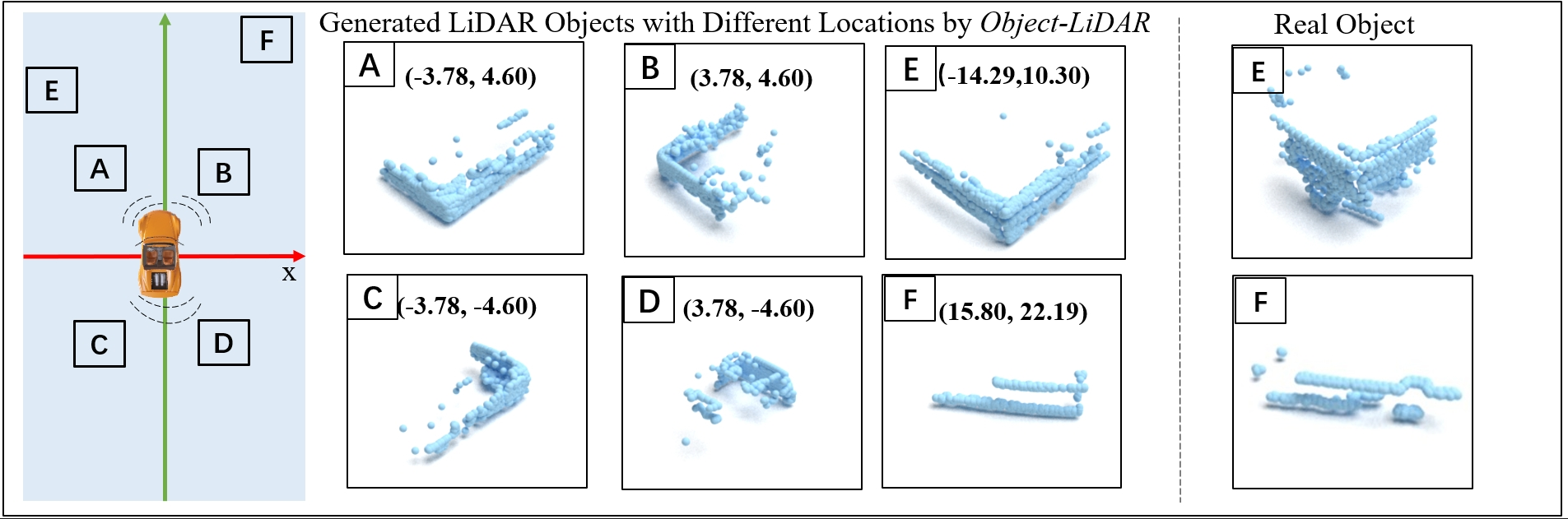}
  \caption{LiDAR scanning patterns vary across different 3D positions, as shown at locations A-F on the map. The first two columns images reveal symmetrical scanning patterns around a vehicle in four quadrants (A, B, C, D). \ourmodel~accurately mirrors real-world LiDAR scans at positions E and F, showcasing its ability to consistently replicate authentic scanning patterns across diverse spatial dimensions.}
  \label{fig:location_3}
\end{figure}
\begin{figure}[t]
  \centering
  \includegraphics[width=1\linewidth]{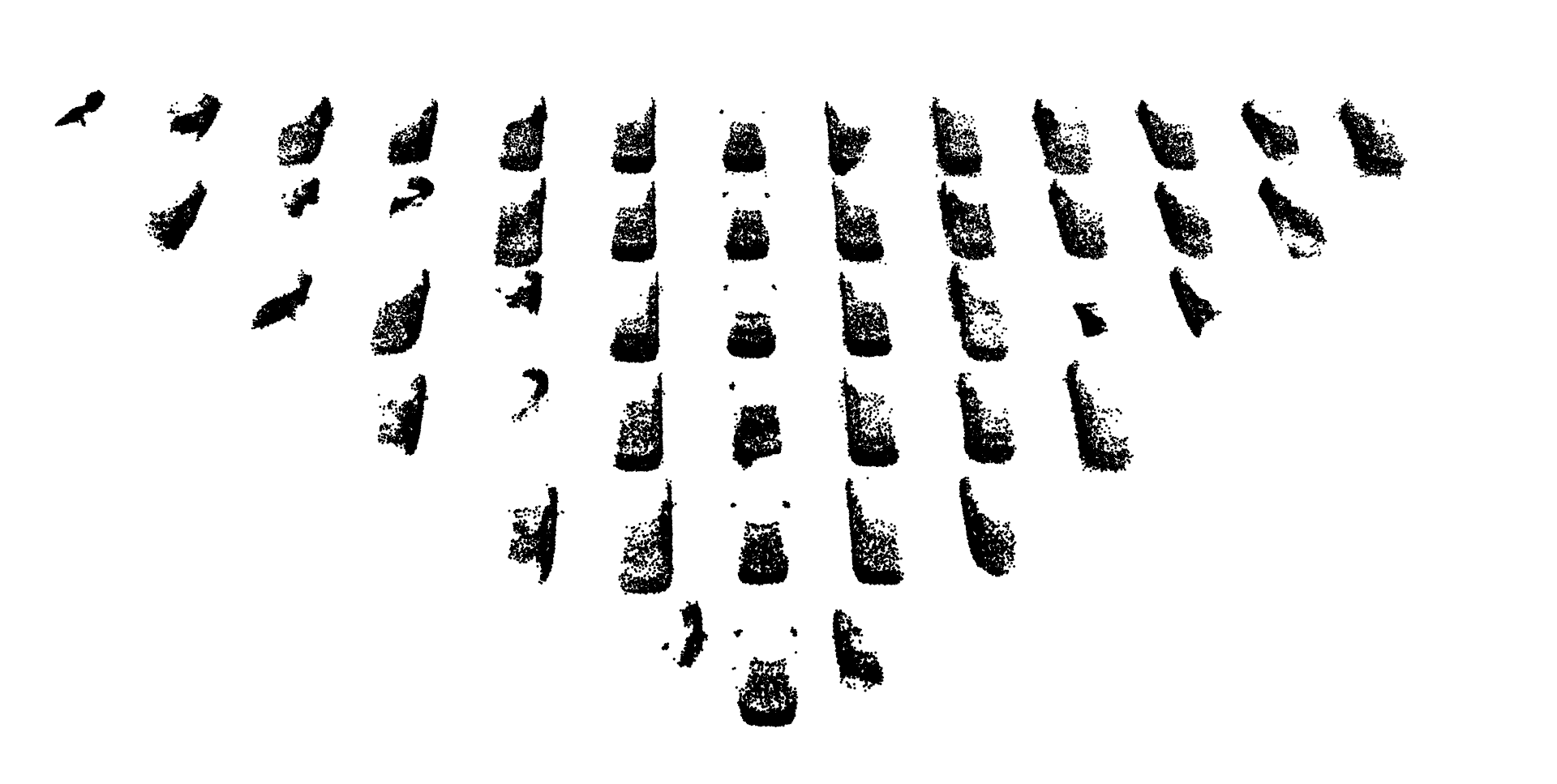}
  \caption{ \textbf{Qualitative results about cars conditioned on various positions.} \ourmodel~controls the generation of LiDAR objects with different 3D location. The farther away an object is, the sparser the point clouds are.}
  \label{fig:location2}
\end{figure}


\section{Discussion about computational efficiency}
\begin{table}[htbp]
  \centering
  \begin{tabular}{c|c|c|c}
    \hline
    Methods & JSD\small{}&Training(h) & Inference(h) \\
    \hline
    LiDARGen[\textcolor{blue}{47}]&    0.67     & 16 h   & 5 h \\
    UltraLiDAR[\textcolor{blue}{37}]&   0.71    & 56 h  & -  \\
    R2DM[\textcolor{blue}{28}]&        0.49     & 12 h & 3 h\\
    \hline
    Ours &   0.42        & 19 h  & 3.5 h \\
  \bottomrule
\end{tabular}
  \caption{\textbf{Comparison of the algorithm efficiency with SOTAs.}}
    \label{tab:freq}
\end{table}
As shown in \cref{tab:freq}, our framework actually achieves comparable training and inference time to SOTA LiDAR diffusion methods. 
Specifically, our inference time is 30\% shorter than LiDARGen~[\textcolor{blue}{47}], yet only 16\% longer than R2DM~[\textcolor{blue}{28}].
Meanwhile, our method significantly surpasses SOTAs in terms of JSD for generation quality, maintaining acceptable computational efficiency. These discussion will be integrated into the final version.
In the future work, we will further optimize the algorithm efficiency.

\section{More Qualitative Results}
\subsection{Conditional LiDAR Completion}

\begin{figure*}[t]
  \centering
  \includegraphics[width=0.9\linewidth]{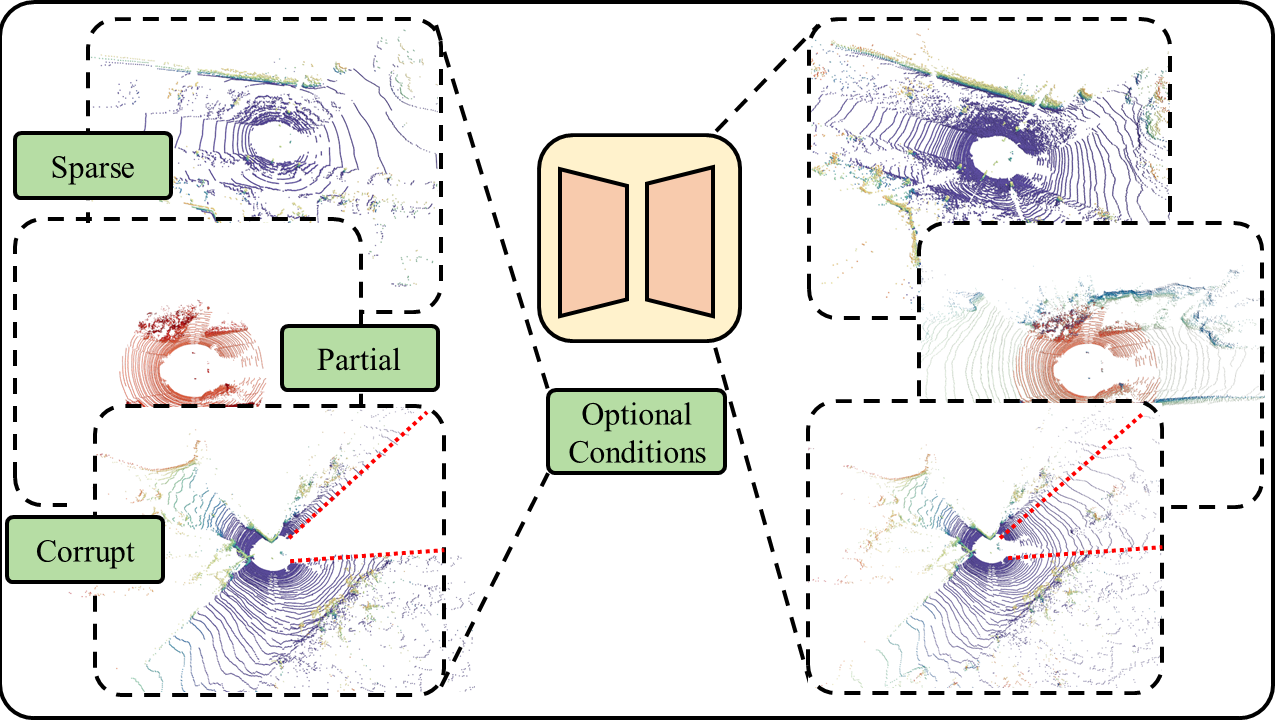}
  \caption{\textbf{\ourmodel~for conditional LiDAR scene generation.} \ourmodel~is capable of handling versatile conditions, achieving realistic and high-fidelity LiDAR data across various conditional generation tasks.}
  \label{fig:condition}
\end{figure*}

For conditional LiDAR generation, LiDARGen and R2DM apply the repaint strategy, replacing parts of the input Gaussian noise with conditions and utilizing the unconditional LiDAR generation network for conditional generation. 
This approach did not engage in further network optimization specifically tailored for the condition-real LiDAR data pairs, impacting the performance on tasks like sparse-to-dense and LiDAR completion. 
In contrast, \ourmodel~handles a wide array of conditions for further network optimization, including scenarios with sparsely distributed laser beams, partial LiDAR point clouds, and even corrupted data (\cref{fig:condition}). This versatility in handling multiple conditions significantly enhances the flexibility and effectiveness of conditional LiDAR generation, better catering to the diverse requirements of generation tasks.

\begin{figure*}[t]
  \centering
  \includegraphics[width=0.8\linewidth]{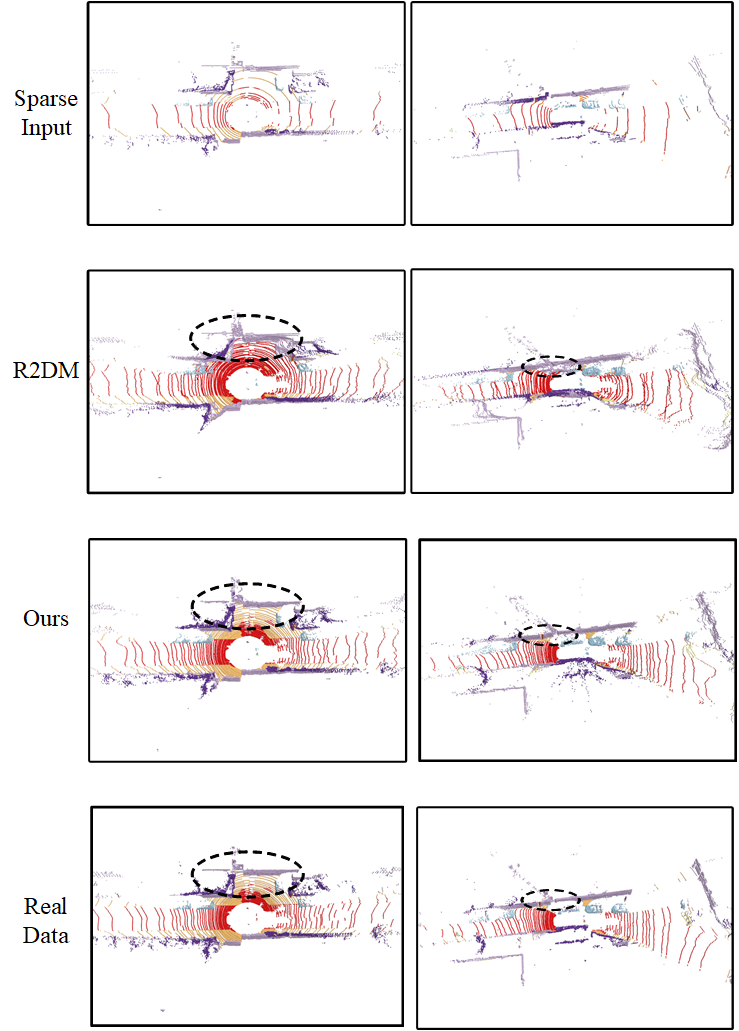}
     \includegraphics[width=0.8\linewidth]{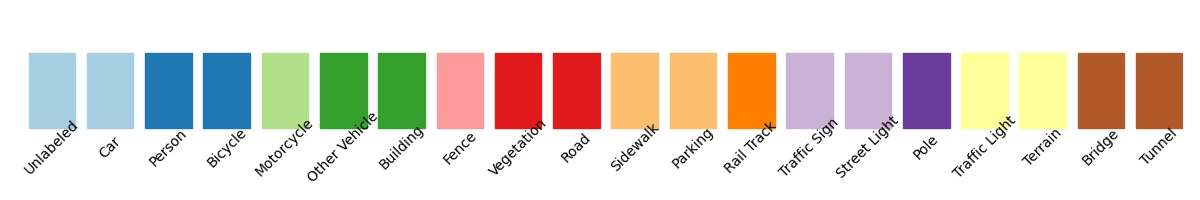}
     \caption{ \textbf{Sparse-to-dense LiDAR Completion.} We use RangeNet-53 to obtain segmentation results.}
  \label{fig:s2d1}
\end{figure*}

\begin{figure*}[t]
  \centering
    \includegraphics[width=0.8\linewidth]{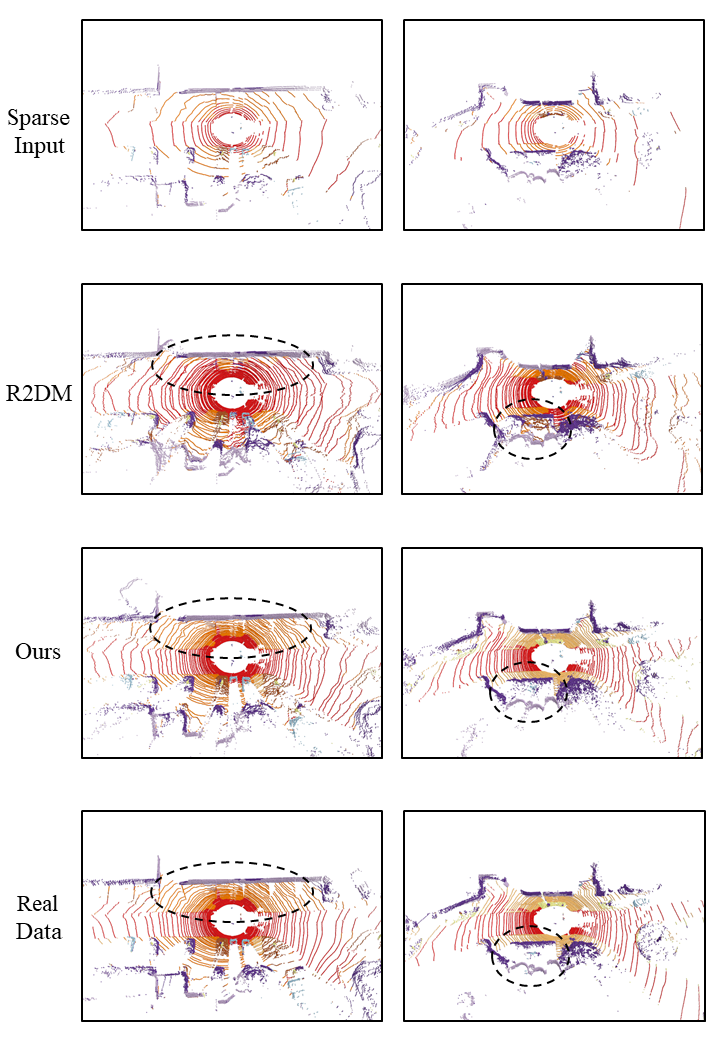}
     \includegraphics[width=0.8\linewidth]{imgs/colormap.jpg}
     \caption{ \textbf{Sparse-to-dense LiDAR Completion.} We use RangeNet-53 to obtain segmentation results.}
  \label{fig:s2d2}
\end{figure*}

\begin{figure*}[t]
  \centering
  \includegraphics[width=0.8\linewidth]{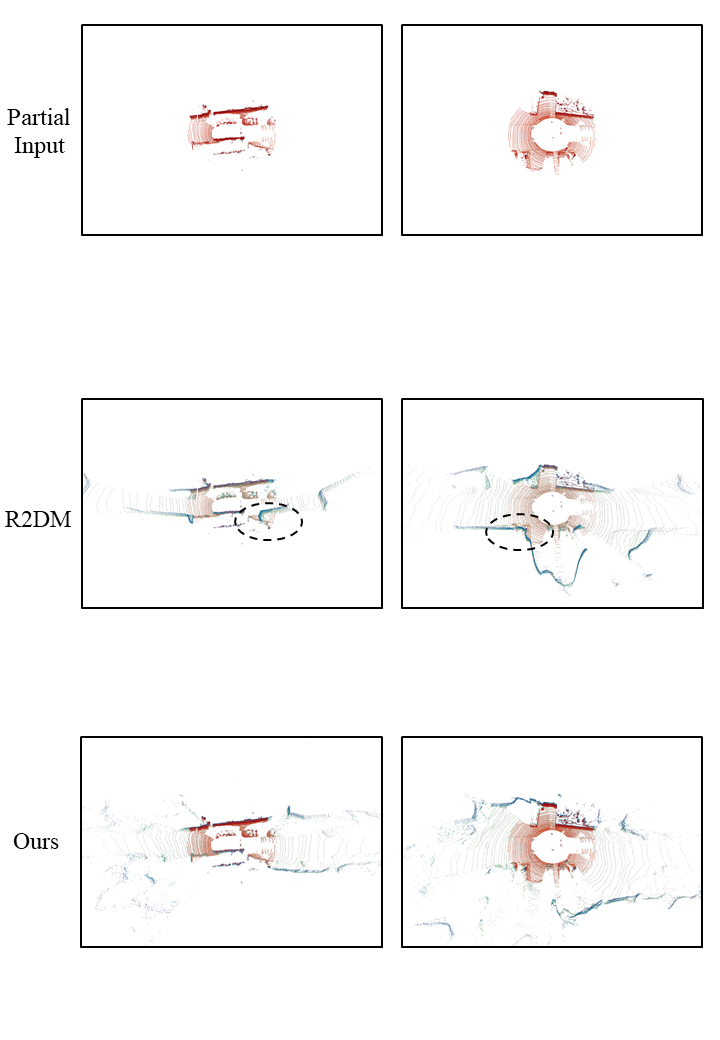}
     \caption{ \textbf{Partial LiDAR Completion.}}
  \label{fig:par1}
\end{figure*}

\begin{figure*}[t]
  \centering
  \includegraphics[width=0.8\linewidth]{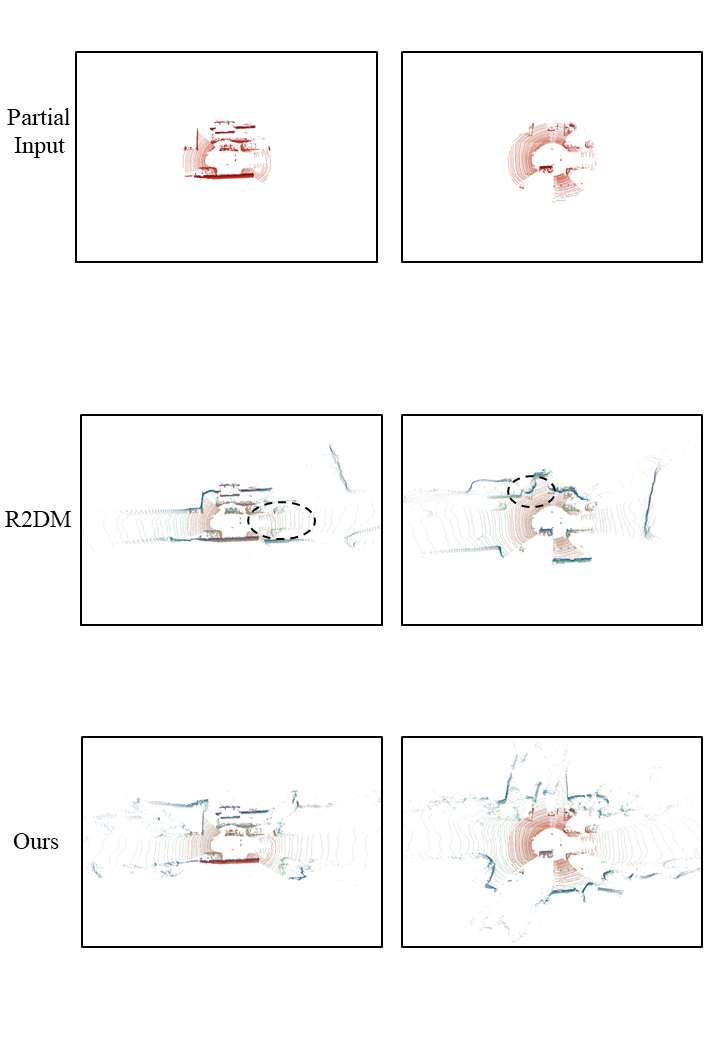}
     \caption{ \textbf{Partial LiDAR Completion.}}
  \label{fig:par2}
\end{figure*}

We show some sparse-to-dense LiDAR completion results in \cref{fig:s2d1} and \cref{fig:s2d2}. The upsampled LiDAR data obtained by \ourmodel~is significantly closer to the real data than R2DM. The segmentation results of road surface and sidewalk on the data synthesized by \ourmodel~are significantly better than those of R2DM.
We also show some partial LiDAR completion results in \cref{fig:par1} and \cref{fig:par2}. The LiDAR data generated by \ourmodel~is significantly more coherent and consistent on both the road surface and the foreground objects.


\subsection{Controllable LiDAR Generation}
\begin{figure*}[t]
  \centering
  \includegraphics[width=1\linewidth]{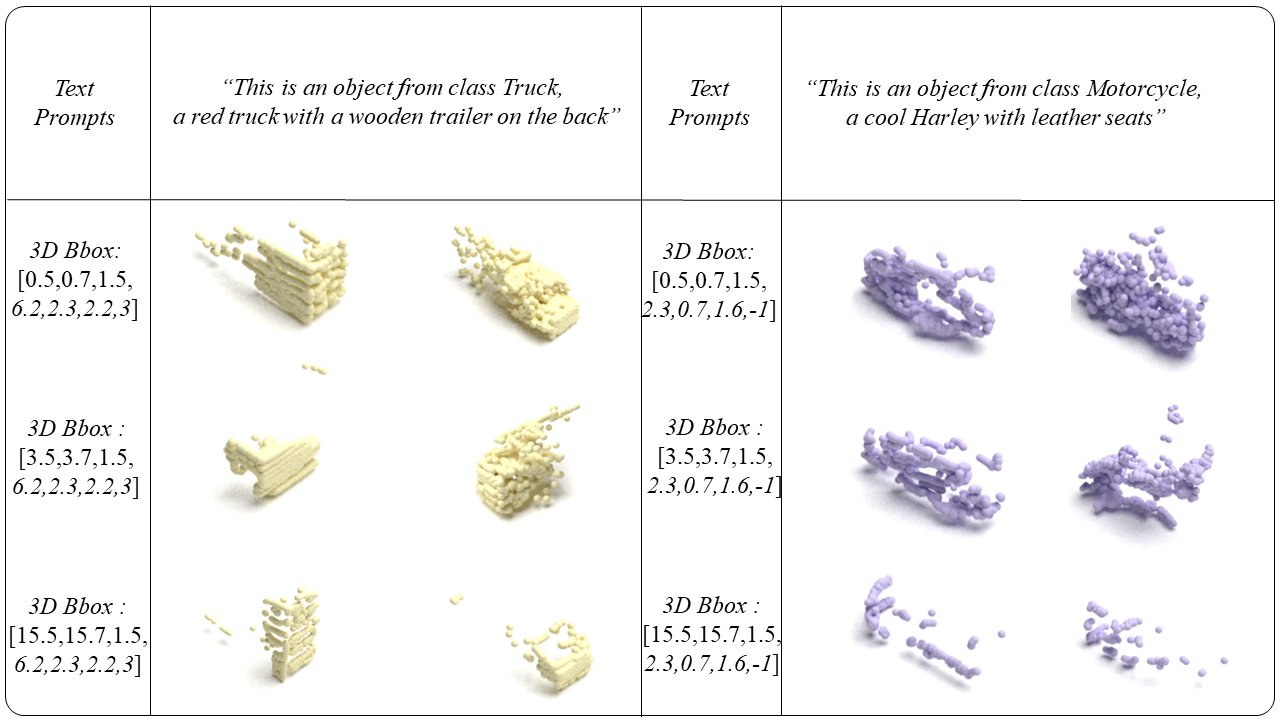}
  \includegraphics[width=1\linewidth]{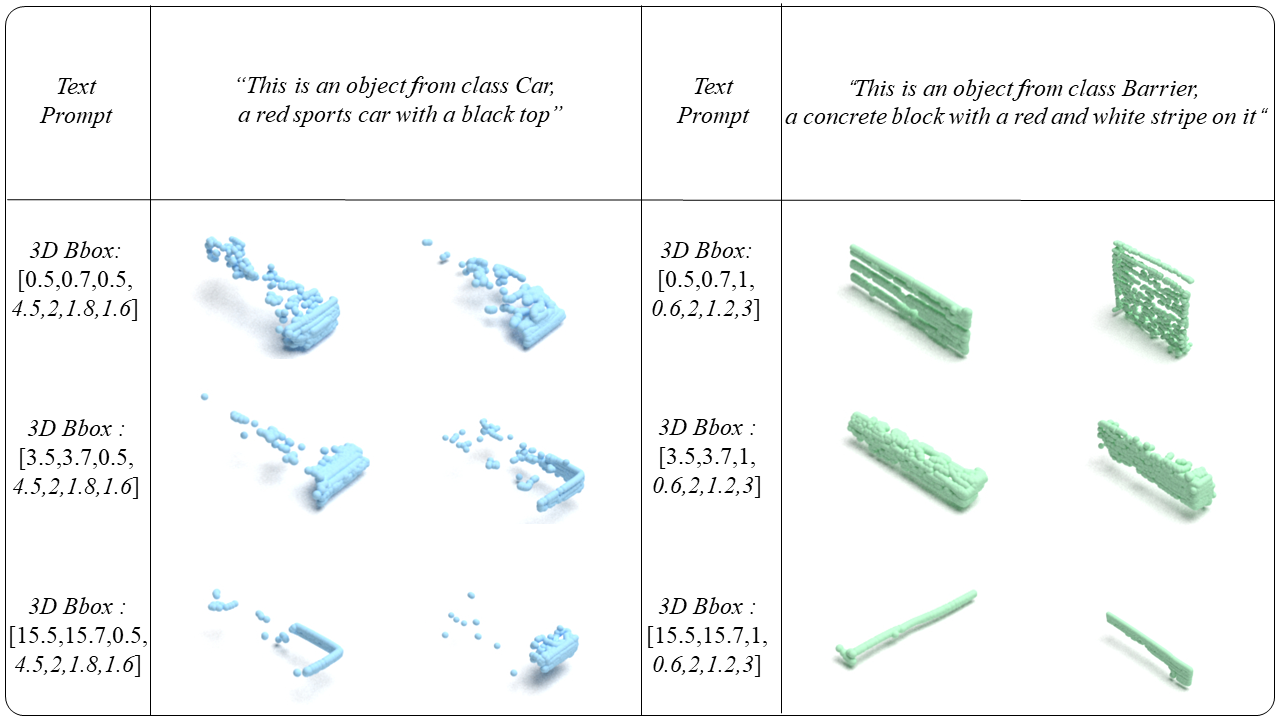}
  \caption{\textbf{Qualitative results about object-level generation on nuScenes.} \ourmodel~controls the generation of LiDAR objects with different 3D location and text prompts. The farther away an object is, the sparser the point clouds are.}
  \label{fig:text_loc}
\end{figure*}

\begin{figure*}[t]
  \centering
  \includegraphics[width=1\linewidth]{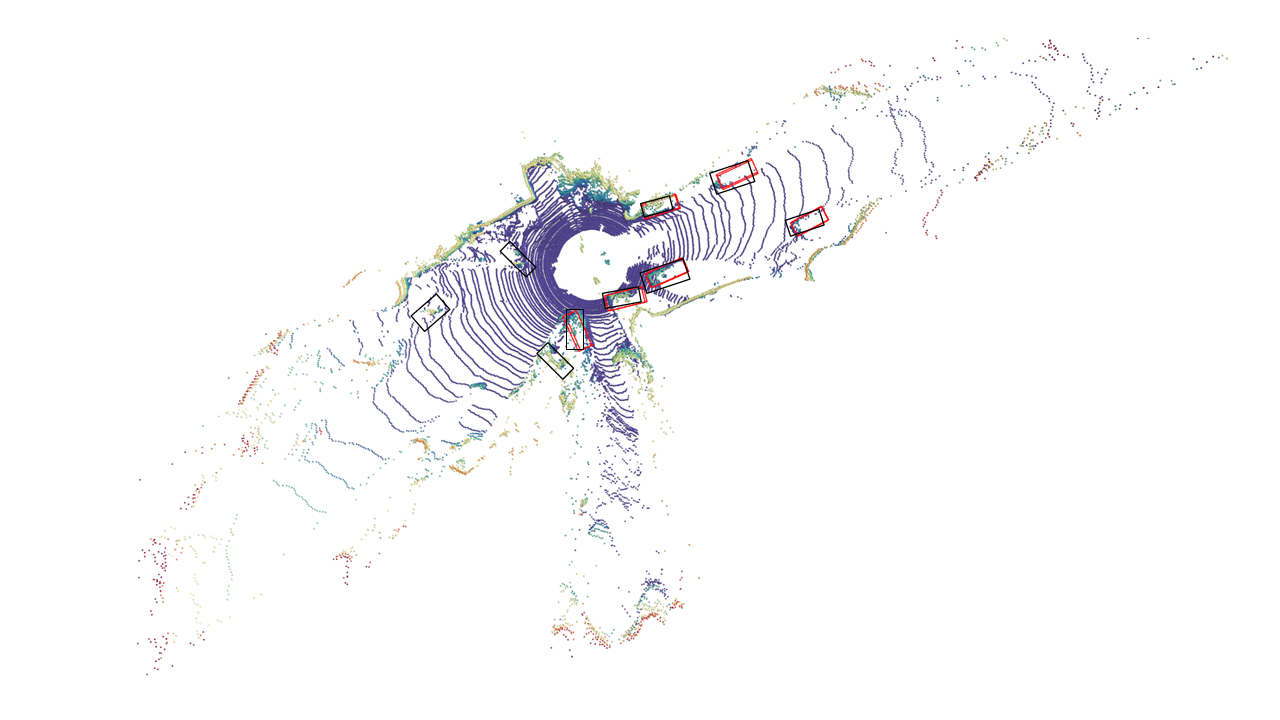}
  \caption{\textbf{Qualitative results about scene-level generation on KITTI-360.} \ourmodel~controls the generation of LiDAR scenes by conditioning foreground objects. \textbf{black} boxes indicate the given object locations as conditions while the \textbf{\textcolor{red}{red}} boxes are the detected ones on the generated LiDAR data.}
  \label{fig:scene_control}
\end{figure*}

The results about controllable object-level generation are shown in \cref{fig:location1}, \cref{fig:location_3}, \cref{fig:location2} and \cref{fig:text_loc}. \ourmodel~controls the generation of LiDAR objects with different 3D location and text prompts. We also show the controllable scene-level generation in \cref{fig:scene_control}, \ourmodel~controls the generation of LiDAR scenes by using foreground objects as conditions.


\vspace{2em}
\section{Enhancing Downstream Task.}
Once trained, \ourmodel~can be readily deployed to generate 3D LiDAR objects in autonomous driving datasets. This approach explicitly addresses the challenges of data scarcity and the long-tail problem in 3D object detection. In this section, we introduce the necessary steps to achieve this goal. 

For a given category, we first generate a textual description that provides the instance-level details. Meanwhile, we simulate the 3D objects with different appearances by flexibly defining the 3D bounding box of the generated sample. It is worth mentioning that \ourmodel~can generate unlimited 3D instances using different combinations of conditions. Afterward, we place these generated 3D objects into the LiDAR point cloud scenes. To avoid oversampling near the ego-vehicle due to the denser points, we opt for uniform sampling $(x_c,y_c)$ across the ground plane, ensuring an even distribution of the objects. In the experiments section, we will show \ourmodel~significantly outperforms the previous method such as GT-Aug.
